\documentclass{article}

\usepackage{microtype}
\usepackage{graphicx}
\usepackage{subcaption}
\usepackage{booktabs} % for professional tables
\usepackage{multirow}

\usepackage{hyperref}

\usepackage[preprint]{icml2026}

\usepackage{amsmath}
\usepackage{amssymb}
\usepackage{mathtools}
\usepackage{amsthm}

\usepackage[capitalize,noabbrev]{cleveref}

\theoremstyle{plain}
\newtheorem{theorem}{Theorem}[section]
\newtheorem{proposition}[theorem]{Proposition}

\theoremstyle{definition}

\theoremstyle{remark}

\usepackage[textsize=tiny]{todonotes}

\icmltitlerunning{}%Submission and Formatting Instructions for ICML 2026

\begin{document}

\twocolumn[
  \icmltitle{
  DRFormer: A Dual-Regularized Bidirectional Transformer \\ 
  for Person Re-identification
  }
%Submission and Formatting Instructions for \\
   % International Conference on Machine Learning (ICML 2026)
  % It is OKAY to include author information, even for blind submissions: the
  % style file will automatically remove it for you unless you've provided
  % the [accepted] option to the icml2026 package.

  % List of affiliations: The first argument should be a (short) identifier you
  % will use later to specify author affiliations Academic affiliations
  % should list Department, University, City, Region, Country Industry
  % affiliations should list Company, City, Region, Country

  % You can specify symbols, otherwise they are numbered in order. Ideally, you
  % should not use this facility. Affiliations will be numbered in order of
  % appearance and this is the preferred way.
  \icmlsetsymbol{equal}{*}

  \begin{icmlauthorlist}
    \icmlauthor{Ying Shu}{bjtu}
    \icmlauthor{Pujian Zhan}{bjtu}
    \icmlauthor{Huiqi Yang}{bjtu}
    \icmlauthor{Hehe Fan}{zju}
    \icmlauthor{Youfang Lin}{bjtu}
    \icmlauthor{Kai Lv}{bjtu}
    %\icmlauthor{}{sch}
    %\icmlauthor{}{sch}
    %\icmlauthor{}{sch}
  \end{icmlauthorlist}

  \icmlaffiliation{bjtu}{Institute of Network Science and Intelligent Systems, Beijing Jiaotong University, Beijing, China}
  \icmlaffiliation{zju}{Zhejiang University, Hangzhou, Zhejiang, China}

  % \icmlaffiliation{comp}{Company Name, Location, Country}
  % \icmlaffiliation{sch}{School of ZZZ, Institute of WWW, Location, Country}

  \icmlcorrespondingauthor{Kai Lv}{lvkai@bjtu.edu.cn}
  % \icmlcorrespondingauthor{Firstname2 Lastname2}{first2.last2@www.uk}

  % You may provide any keywords that you find helpful for describing your
  % paper; these are used to populate the "keywords" metadata in the PDF but
  % will not be shown in the document
  \icmlkeywords{Machine Learning, ICML}

  \vskip 0.3in
]

% this must go after the closing bracket ] following \twocolumn[ ...

% This command actually creates the footnote in the first column listing the
% affiliations and the copyright notice. The command takes one argument, which
% is text to display at the start of the footnote. The \icmlEqualContribution
% command is standard text for equal contribution. Remove it (just {}) if you
% do not need this facility.

% Use ONE of the following lines. DO NOT remove the command.
% If you have no special notice, KEEP empty braces:
\printAffiliationsAndNotice{}  % no special notice (required even if empty)
% Or, if applicable, use the standard equal contribution text:
% \printAffiliationsAndNotice{\icmlEqualContribution}

\begin{abstract}

Both fine-grained discriminative details and global semantic features can contribute to solving person re-identification challenges, such as occlusion and pose variations. 
Vision foundation models (\textit{e.g.}, DINO) excel at mining local textures, and vision-language models (\textit{e.g.}, CLIP) capture strong global semantic difference. 
Existing methods predominantly rely on a single paradigm, neglecting the potential benefits of their integration. 
In this paper, we analyze the complementary roles of these two architectures and propose a framework to synergize their strengths by a \textbf{D}ual-\textbf{R}egularized Bidirectional \textbf{Transformer} (\textbf{DRFormer}). 
The dual-regularization mechanism ensures diverse feature extraction and achieves a better balance in the contributions of the two models.
Extensive experiments on five benchmarks show that our method effectively harmonizes local and global representations, achieving competitive performance against state-of-the-art methods.

\end{abstract}

\section{Introduction}

%异质性heterogeneity auxiliary regularizers unimodal bias regularization
Person re-identification (ReID) aims to retrieve a target pedestrian from the gallery given a query image. The ReID task is highly challenging due to blur, occlusions, illumination changes, and huge pose variations. 
We expect the model to capture holistic human body structures, \underline{\textbf{and}} to attend to subtle differences between similar individuals, such as accessories and clothing textures.

\begin{figure}[ht]
  \vskip 0.2in
  \begin{center}
  \centerline{\includegraphics[width=1\columnwidth]{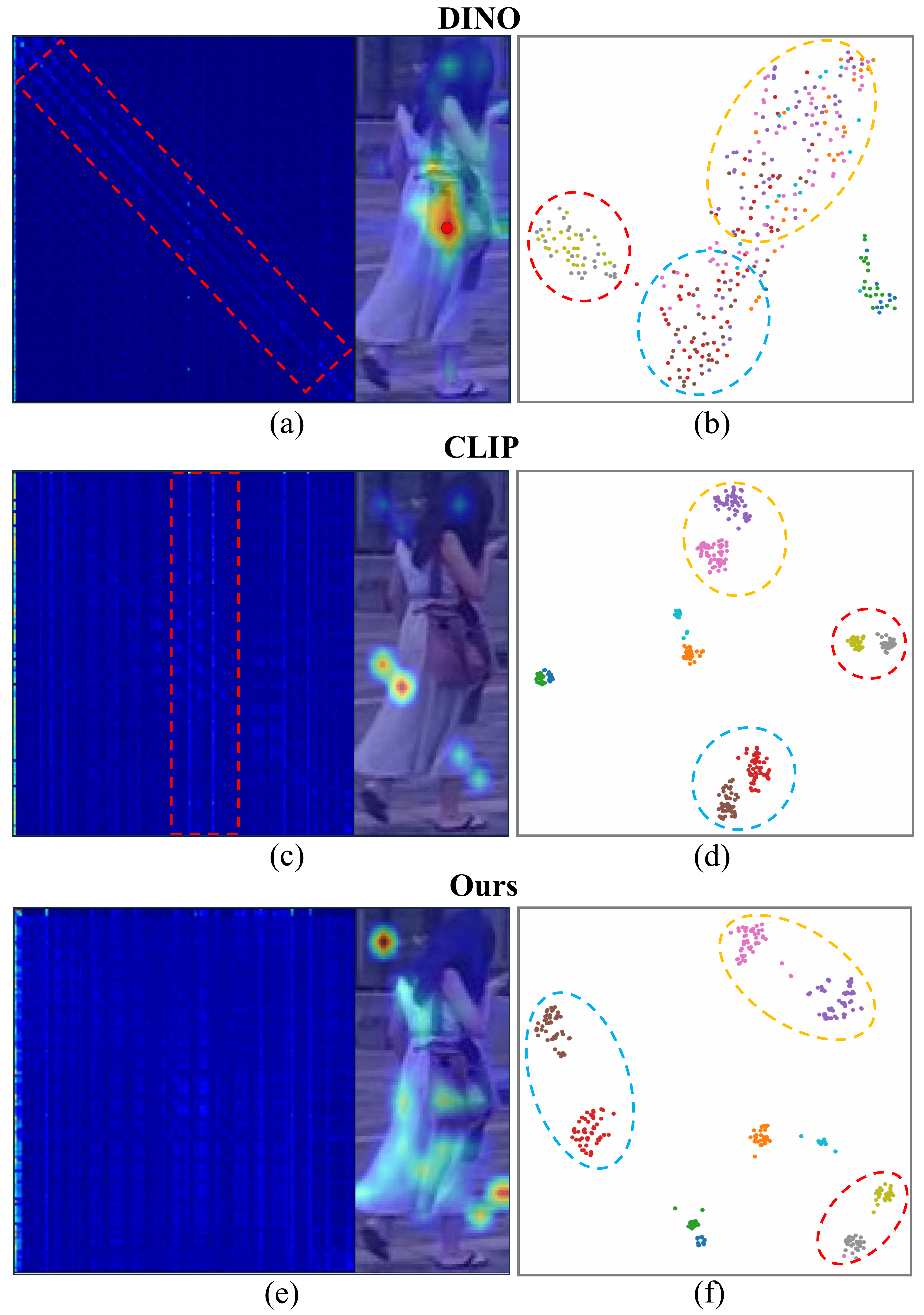}}
    \caption{Comparison of attention patterns and feature distributions for DINO, CLIP and our method. The left column visualizes attention weights for all tokens and a selected token: Figure (a) shows the query image token at the red dot, while Figures (c) and (e) depict attention weights of the [CLS] token. The right column presents feature distributions of fine-tuned DINO, fine-tuned CLIP, and our method on some samples.
}
    \label{intro}
  \end{center}
\end{figure}

Many existing ReID methods focus on discriminative fine-grained details \cite{DBLP:conf/eccv/ZhuGYZWT22, DBLP:journals/mva/HuWL25} \underline{\textbf{or}} global semantic features \cite{DBLP:conf/aaai/LiSL23, DBLP:conf/cvpr/YangWW0G024, DBLP:journals/corr/abs-2507-10203}. 
On the one hand, vision foundation models (VFMs) like DINO \cite{DBLP:conf/iccv/CaronTMJMBJ21, DBLP:journals/tmlr/OquabDMVSKFHMEA24} have demonstrated strong capability in fine-grained feature mining.  
For example, a catastrophic forgetting score is employed to measure the gap between the pre-training and fine-tuning data of DINO, followed by the design of a ReID-specific module \cite{DBLP:journals/corr/abs-2111-12084}.
On the other hand, vision–language models (VLMs) like CLIP \cite{DBLP:journals/corr/abs-2103-00020} excel at extracting high-level semantic features. 
CLIP-ReID \cite{DBLP:conf/aaai/LiSL23} applies CLIP to person ReID by fully exploiting its cross-modal describing ability. 
However, in ReID filed, most existing methods rely predominantly on one of the two perspectives, neglecting the potential benefits of their integration.
The potential for concurrently exploiting the local fine-grained discriminative power of DINO and the global semantic understanding of CLIP remains insufficiently explored in the current literature.

In this paper, we first investigate a question: \textbf{what specific roles DINO and CLIP play in person ReID?} 
1) As shown in Figure \ref{intro} (a) and (c), we observe a clear distinction: DINO demonstrates a strong capability to attend to fine-grained local features, with attention peaks appearing along the diagonal; whereas CLIP only concentrates attention on a few fixed tokens and lacks local discriminability. 
2) Figures \ref{intro} (b) and (d) illustrate the feature distributions of fine-tuned DINO and CLIP on pedestrian identities. We find that DINO features are highly overlapping and scattered, while CLIP features exhibit a clustered structure. 
The reason is that global semantic context can benefit inter-class separation for person ReID. 
The above comparisons reveal a compelling opportunity: DINO provides the necessary fine-grained texture, while CLIP ensures the essential semantic separability. Thus, the key to improving ReID performance lies in harmonizing these distinct yet complementary strengths.

% naive strategy involves direct feature fusion (e.g., concatenation or element-wise addition)
% To achieve a better fus. 
% First, we need to enhance the feature extraction ability of each individual model, because high-quality representations are the foundation for effective fusion. Second, we need to ensure efficient collaboration between the two models, so that the model can fully integrate their complementary advantages.

Motivated by this insight, we address the pivotal question: \textbf{how to effectively combine the capabilities of DINO and CLIP?}
To this end, we design our framework around two critical principles:
For each individual model, we aim to enhance feature diversity, positing that more diverse information serves as a richer substrate for effective fusion. 
For the fusion of the two models, we prioritize effective collaboration to fully integrate their complementary advantages. 

%\textcolor{red}{ensuring that both fine-grained details and global semantic features are equally utilized.}
 
%To this end, we design our framework around two critical principles:
%Intra-model Level: we aim to enhance feature diversity, positing that more diverse information serves as a richer substrate for effective fusion. 
%Inter-model Level: we should ensure effective collaboration to fully integrate their complementary advantages.

%In this paper, we present DRFormer, a \textbf{D}ual-\textbf{R}egularized Bidirectional \textbf{Transformer} \textcolor{red}{that employs bidirectional cross-attention to alternate DINO and CLIP features as queries.}
%We further introduce two regularizers: one for intra-model token diversity, \textcolor{red}{which maximizes cosine distances to promote complementary feature learning}, and another for inter-model bias, which balances model contributions to eliminate learning suppression.
%Guided by our theoretical analysis, we employ a strategy that reduces the prediction error of each individual model.
%Our contributions can be summarized as:
In this paper, we propose the \textbf{D}ual-\textbf{R}egularized Bidirectional \textbf{Transformer} (\textbf{DRFormer}).
We employ a bidirectional cross-attention mechanism where features from DINO and CLIP alternately serve as queries to facilitate deep interaction.
Furthermore, we introduce two auxiliary regularizers.
The intra-model token diversity regularizer aims to promote diverse feature learning of learnable tokens. 
%We maximize the cosine distances of these tokens to encourage them to focus on complementary regions. 
We maximize the cosine distances of these tokens to drive the model to attend to complementary regions.
The inter-model bias regularizer aims to eliminate inter-model learning suppression and achieve a better contribution ratio of the two models.
Guided by our theoretical analysis, we formulate this regularizer to minimize the prediction error of each individual model.
Our contributions can be summarized as:

\begin{itemize}
  \item For the \textit{what} question, we analyze and validate the complementary roles of DINO and CLIP: DINO captures local discriminative patterns essential for fine-grained classification, while CLIP imposes global semantic constraints to enhance identity separability.
  \item For the \textit{how} question, we propose a bidirectional transformer module and introduce two regularizers. The intra-model regularizer encourages fully exploiting diverse representations, and the inter-model regularizer balances the contributions of DINO and CLIP.
  \item On five person ReID datasets, our model achieves competitive performance compared with state-of-the-art ReID methods, demonstrating the effectiveness of the proposed approach.
\end{itemize}

%%%%%%%%%%%%%%%%%%%%%%%%%%%%%%%%%%%%%%%%%%%%%%%%%%%%%%%%%%%%%%%%

\section{Related Work}
\subsection{Person Re-identification}

Person re-identification aims to match pedestrians across non-overlapping camera views, identifying the same identity despite pose variations and occlusions. The research landscape has evolved from CNN-based architectures to Transformer-based approaches.
CNN-based methods \cite{DBLP:conf/eccv/SunZYTW18, DBLP:conf/mm/HeLLLCM23} primarily focus on learning discriminative feature representations via metric learning. 
Nowadays, Transformer-based methods \cite{DBLP:conf/iccv/He0WW0021, DBLP:conf/cvpr/ZhuKLLTS22} are proposed and leverage self-attention to model long-range dependencies. The pioneering work TransReID \cite{DBLP:conf/iccv/He0WW0021} introduces jigsaw patches and side information embeddings to enhance feature robustness.
In this paper, we propose a Transformer-based framework to achieve robust and discriminative person re-identification.

\subsection{Large-scale Pre-training Models}
Vision foundation models (VFMs) \cite{DBLP:conf/cvpr/HeCXLDG22, DBLP:conf/iccv/KirillovMRMRGXW23, DBLP:journals/tmlr/OquabDMVSKFHMEA24} and vision–language models (VLMs) \cite{DBLP:journals/corr/abs-2103-00020, DBLP:conf/icml/JiaYXCPPLSLD21, DBLP:conf/iccv/ZhaiM0B23} are pre-trained on large-scale datasets and show strong image understanding capabilities. VFMs typically act as visual backbones and are widely used in many downstream tasks \cite{DBLP:conf/cvpr/Li0XL0NS23, DBLP:journals/corr/abs-2508-11256}.
DINO \cite{DBLP:conf/iccv/CaronTMJMBJ21, DBLP:journals/tmlr/OquabDMVSKFHMEA24} simplifies self-supervised training by directly predicting the output of a teacher network and excels at capturing fine-grained details. MAE \cite{DBLP:conf/cvpr/HeCXLDG22} significantly improves masked image modeling (MIM) for vision transformers \cite{DBLP:conf/iclr/DosovitskiyB0WZ21}. 
Meanwhile, VLMs show strong generalization across diverse tasks. CLIP \cite{DBLP:journals/corr/abs-2103-00020} aligns images and texts in a shared embedding space via contrastive learning, enabling effective capture of global semantic context. SigLIP \cite{DBLP:conf/iccv/ZhaiM0B23} further enhances this by adopting a sigmoid-based objective.

Many ReID methods are built upon existing VFMs or VLMs. Based on DINO, PersonViT \cite{DBLP:journals/mva/HuWL25} combines MIM with discriminative contrastive learning to extract global and local features. CLIP-ReID \cite{DBLP:conf/aaai/LiSL23} uses prompt learning to generate text tokens for each identity and fully leverages the prior knowledge of CLIP. PromptSG \cite{DBLP:conf/cvpr/YangWW0G024} further improves generalization by converting image features into pseudo tokens. Different from these works that focus on only one aspect, our method aims to combine the strengths of VFMs and VLMs to achieve better performance.

\subsection{Feature Fusion Methods}
Feature fusion integrates information from different levels to obtain stronger representations. The simplest approach is direct concatenation, which requires careful alignment of feature map dimensions. With the introduction of transformers, many attention-based feature fusion methods \cite{DBLP:conf/cvpr/Xu00WO18, DBLP:journals/tnn/ZhouZLJ22} have been proposed. Dual attention network (DANet) is proposed to adaptively integrate local features with their global dependencies \cite{DBLP:conf/cvpr/FuLT0BFL19}. GCNet \cite{DBLP:conf/iccvw/0001XLWH19} enhances local features using global contextual information, while significantly reducing computational complexity. For feature fusion of pre-trained models, it is more challenging due to the inherent feature-level mismatch. In other fields such as monocular depth estimation (MDE), some work \cite{DBLP:journals/corr/abs-2510-09320} guides the alignment of DINO and CLIP through text. DeCLIP \cite{DBLP:journals/corr/abs-2508-11256} enhances the discriminability and spatial consistency of CLIP’s dense features by distilling attention maps of DINO, which requires training with large-scale datasets. In this paper, we achieve efficient fusion of DINO and CLIP features through a bidirectional transformer with the dual-regularization mechanism.

%It is very challenging to build a lightweight module that efficiently fuses DINO and CLIP features and can also achieve outstanding performance on small-scale person ReID datasets.

\section{Proposed Method}

% The overall framework, shown in Figure \ref{method}, comprises a \textcolor{red}{DINO} \cite{DBLP:journals/tmlr/OquabDMVSKFHMEA24} encoder, a \textcolor{red}{CLIP} \cite{DBLP:journals/corr/abs-2103-00020} image encoder, and a bidirectional fusion transformer. 
% Through some \textcolor{red}{toy} experiments, we identify two key challenges in the feature fusion process. 
% %这里的toy experiment通常有过于简单的意思
% %我感觉下列替代会更好
% %Based on our empirical analysis
% %Through preliminary experiments
% Accordingly, we adopt a bidirectional transformer to fuse DINO and CLIP features and introduce intra- and inter-model regularizers to address the challenges. The entire framework is trained end-to-end.

\subsection{Key Observations and Motivation}

Before answering the question of \textit{how} to integrate the capabilities of DINO and CLIP, we first conduct experiments to identify the key challenges in the feature fusion process. As shown in Figure \ref{observation}, the challenges lie in two aspects: 

\begin{figure}[ht]
  \vskip 0.2in
  \begin{center}
  \centerline{\includegraphics[width=\columnwidth]{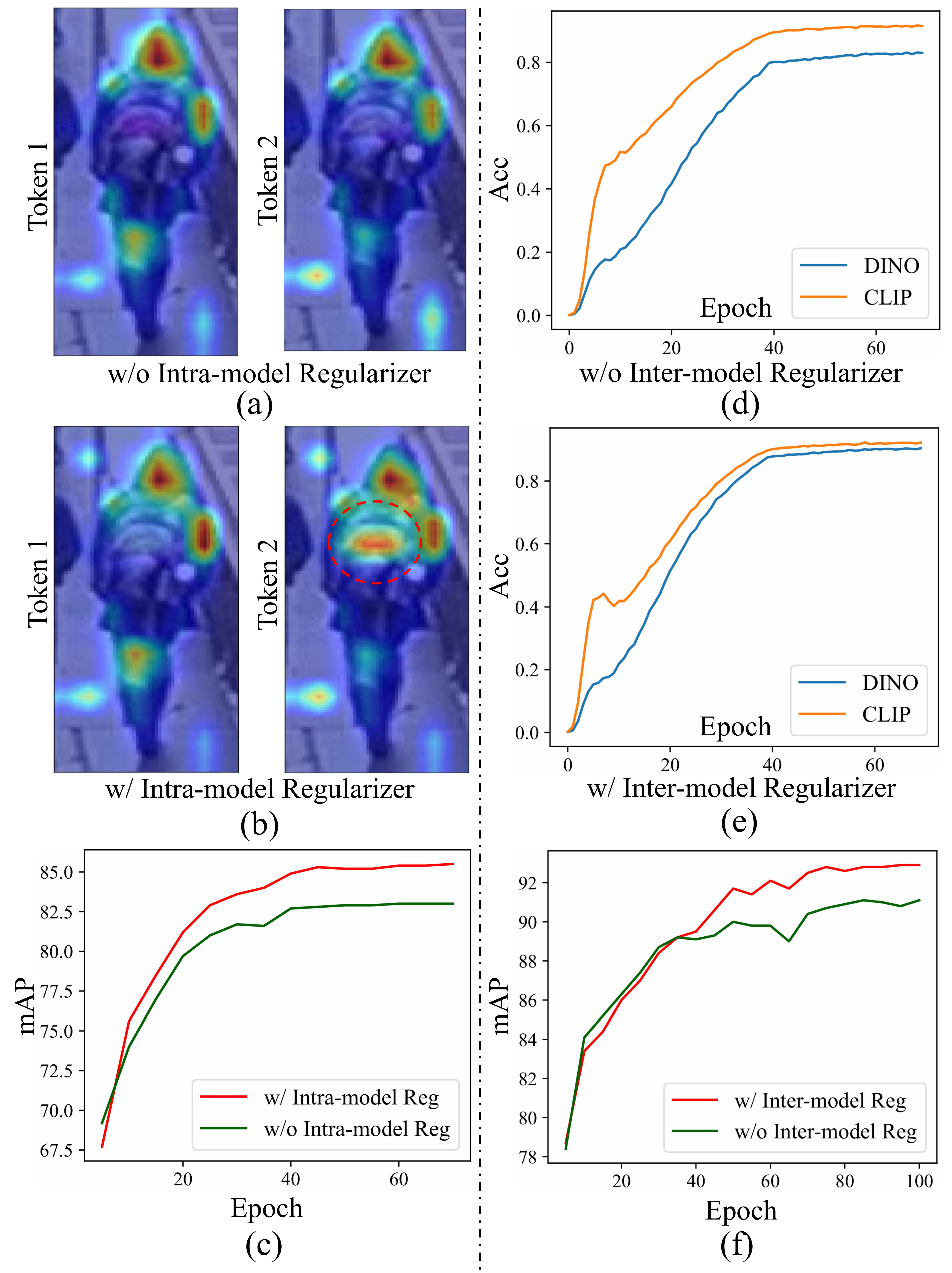}}
    \caption{The left and right columns illustrate the motivation and effects of the intra-model and inter-model regularizers, respectively. Left: Figures (a) and (b) depict the attention weights of the two learnable tokens before and after the intra-model regularizer on DukeMTMC dataset. Figure (c) shows the performance gain (2.5\% $\uparrow$). Right: Figures (d) and (e) present the DINO and CLIP accuracy curves before and after adding the inter-model regularizer on Market-1501 dataset. Figure (f) shows the performance improvement (1.8\% $\uparrow$).
    }
    \label{observation}
  \end{center}
\end{figure}

\begin{figure*}[ht]
  \vskip 0.2in
  \begin{center}
  \centerline{\includegraphics[width=2\columnwidth]{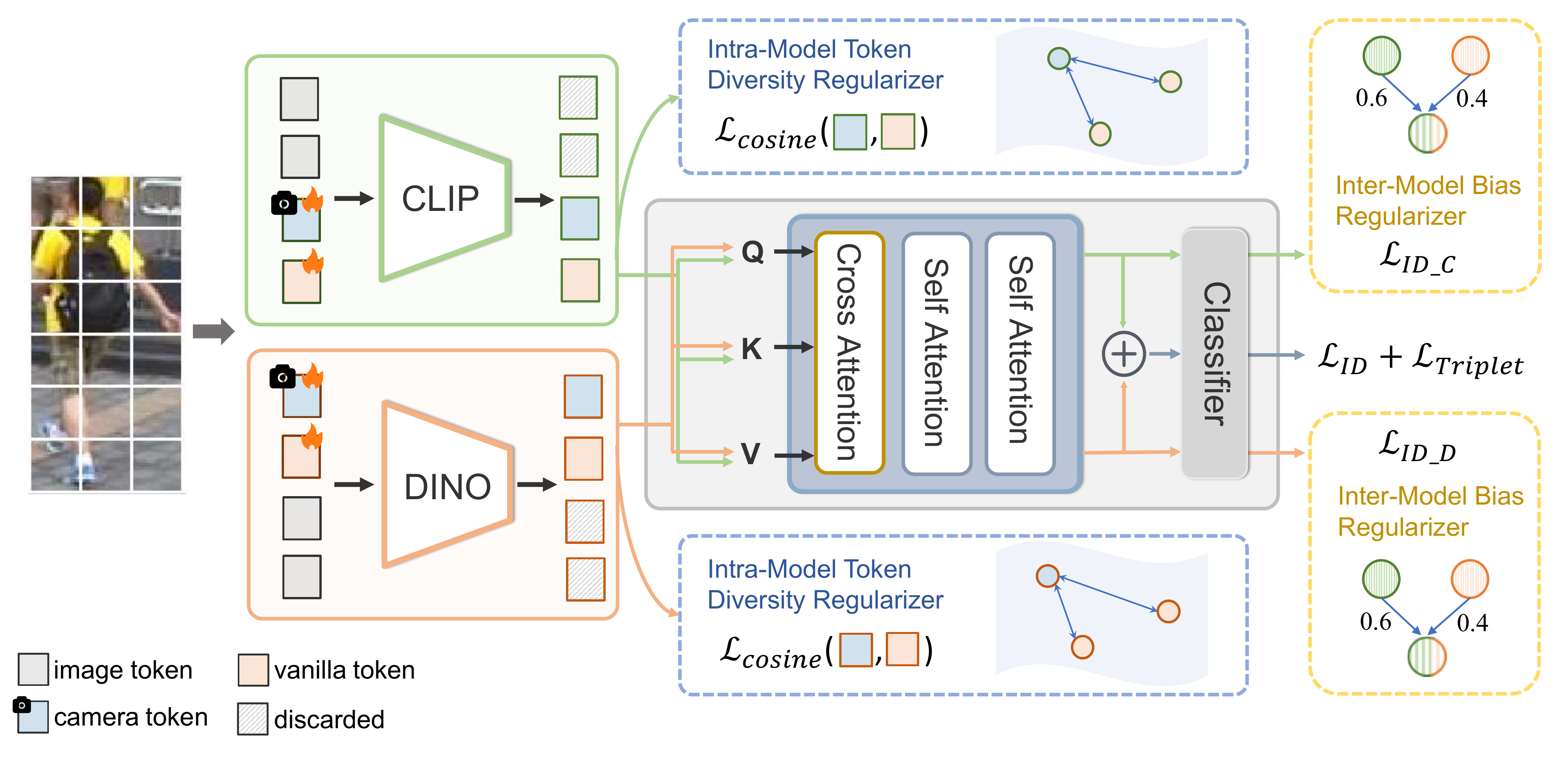}}
    \caption{The framework of our method. DRFormer mainly consists of a CLIP image encoder, a DINO encoder, and a bidirectional fusion transformer. An input image is split into patches, which, along with learnable tokens, are fed separately into CLIP and DINO. The outputs of the learnable tokens from the two models serve two purposes: performing intra-model regularization and acting as queries for cross-attention. The two outputs of the fusion module are concatenated and fed into the linear classifier. Additionally, each output is also individually passed to the classifier for inter-model regularization.}
    \label{method}
  \end{center}
\end{figure*}

%In our experiments, we find that simply fusing DINO and CLIP features via a bidirectional transformer for classification does not yield satisfactory performance. 
\textbf{Insufficient Feature Exploitation within DINO and CLIP.} In order to fully extract useful information from DINO and CLIP, we argue that relying solely on the classification loss is not enough. As shown in Figure \ref{observation} (a), we observe that the attention regions of different learnable tokens are almost identical. This means that different learnable tokens within the same model tend to extract relatively homogeneous features. 
% Additionally, both tokens do not pay attention to the backpack, which can sometimes serve as crucial cues for discrimination. 
Moreover, both tokens fail to attend to the backpack region, which can provide important cues for distinguishing similar individuals. These observations suggest that additional constraints are required to encourage diverse and complementary feature extraction within each model.

To encourage the tokens to extract diverse features, we introduce the \textbf{intra-model token diversity regularizer}. 
As shown in Figure \ref{observation} (b), the attention map of token 2 highlights the backpack region, illustrating the diversity of the learned features.
Figure \ref{observation} (c) shows that the regularizer yields a performance gain of 2.5\% on DukeMTMC, validating its effectiveness. 

\textbf{Imbalanced Learning of DINO and CLIP.} Due to the different pre-training paradigms and datasets, the features extracted by DINO and CLIP are substantially different. 
Although DINO and CLIP are configured symmetrically, we hypothesize that their learning processes are inherently imbalanced. To verify this, we jointly trained both models within a unified framework, yet evaluated the performance of each branch independently on Market-1501. 
As shown in Figure \ref{observation} (d), the accuracy of CLIP increases significantly faster than that of DINO at the beginning. As the training process approaches convergence, a stable and parallel gap emerges. The experimental results reveal that the DINO and CLIP encoders exhibit different levels of learning difficulty. 

% Thus, additional supervision is required to encourage both models to fully exploit their capabilities. 
% Considering that different datasets have different characteristics, it is necessary to control individual contributions of DINO and CLIP to combine their respective strengths \cite{DBLP:journals/corr/abs-2507-10203}. 
To harmonize their individual contributions, we introduce the \textbf{inter-model bias regularizer}. Figure \ref{observation} (e) and (f) presents the experimental results when incorporating the inter-model regularizer. The regularizer effectively eliminates the accuracy gap between DINO and CLIP, and improves performance by a large margin on Market-1501.
%For example, DukeMTMC involves more variations in camera viewpoints and illumination compared to Market-1501.

\subsection{Bidirectional Fusion Transformer}
% Our method aims to effectively leverage the complementary strengths of DINO and CLIP. DINO excels at fine-grained detail extraction, and CLIP captures global semantic context. 
% Given an input image, we employ a DINO encoder $\mathcal{E}_{\text{DINO}}$ and a CLIP image encoder $\mathcal{E}_{\text{CLIP}}$ to extract complementary features simultaneously.
% Specifically, we divide the image into $M$ non-overlapping patches and embed them into patch-token sequences, denoted as $\boldsymbol{I}^D_{1 \sim M} \in \mathbb{R}^{M \times d_D}$ for DINO and $\boldsymbol{I}^C_{1 \sim M} \in \mathbb{R}^{M \times d_C}$ for CLIP, where $d_D$ and $d_C$ are the token embedding dimensions of the two encoders. 
% In addition, we introduce $N$ learnable tokens (\textit{i.e.}, $\boldsymbol{T}^D_{1 \sim N} \in \mathbb{R}^{N \times d_D} $ and $\boldsymbol{T}^C_{1 \sim N} \in \mathbb{R}^{N \times d_C}$) for each branch, and feed $(\boldsymbol{I}^D,\boldsymbol{T}^D)$ and $(\boldsymbol{I}^C,\boldsymbol{T}^C)$ into $\mathcal{E}_{\text{DINO}}$ and $\mathcal{E}_{\text{CLIP}}$, respectively.
% The learnable tokens are used as queries to extract useful information from two models for feature fusion. 
% We discard the outputs of image tokens (\textit{i.e.}, $\boldsymbol{f}^D_{I}$ and $\boldsymbol{f}^C_{I}$) to reduce the computational cost. Only the output features corresponding to the learnable tokens, \textit{i.e.}, $\boldsymbol{f}^D_{T}$ and $\boldsymbol{f}^C_{T}$ are fed into the fusion module.

Our method aims to effectively leverage the complementary strengths of DINO and CLIP. DINO excels at fine-grained detail extraction, and CLIP captures global semantic context. We use a DINO encoder $\mathcal{E}_{\text{DINO}}$ and a CLIP image encoder $\mathcal{E}_{\text{CLIP}}$ to extract complementary features from the same image simultaneously. Specifically, we divide the image into $M$ non-overlapping patches and embed them into patch-token sequences, denoted as $\boldsymbol{I}^D_{1 \sim M} \in \mathbb{R}^{M \times d_D}$ for DINO and $\boldsymbol{I}^C_{1 \sim M} \in \mathbb{R}^{M \times d_C}$ for CLIP, where $d_D$ and $d_C$ are the token embedding dimensions of the two encoders. 
We introduce $N$ learnable tokens (\textit{i.e.}, $\boldsymbol{T}^D_{1 \sim N} \in \mathbb{R}^{N \times d_D} $ and $\boldsymbol{T}^C_{1 \sim N} \in \mathbb{R}^{N \times d_C}$) for each model. These tokens are concatenated with image tokens and then fed into two models, respectively, as shown in Equation \ref{2-1} and \ref{2-2}.

% We use a DINO encoder $\mathcal{E}_{\text{DINO}}$ and a CLIP image encoder $\mathcal{E}_{\text{CLIP}}$ to extract complementary features from the same image simultaneously. 
% An image is divided into $M$ non-overlapping patches, which are embedded as image tokens, \textit{i.e.}, $\boldsymbol{I}^D_{1 \sim M} \in \mathbb{R}^{M \times d_D}$ for DINO and $\boldsymbol{I}^C_{1 \sim M} \in \mathbb{R}^{M \times d_C}$ for CLIP. 
%These image tokens are fed into two models, respectively, together with $N$ learnable tokens $\boldsymbol{T}^D_{1 \sim N} \in \mathbb{R}^{N \times d_D} $ and $\boldsymbol{T}^C_{1 \sim N} \in \mathbb{R}^{N \times d_C}$. 
% The learnable tokens are used as queries to extract useful information from two models for feature fusion. 
% We discard the output of image tokens to reduce the computational cost. Only the output features corresponding to the learnable tokens, \textit{i.e.}, $\boldsymbol{f}^D_{T}$ and $\boldsymbol{f}^C_{T}$ are fed into the fusion module. 

\begin{align}
    \label{2-1}
    [\boldsymbol{f}^D_{T};\boldsymbol{f}^D_{I}]&= \mathcal{E}_{\text{DINO}}([\boldsymbol{T}^D_1,...,\boldsymbol{T}^D_N; \boldsymbol{I}^D_{1}, ..., \boldsymbol{I}^D_{M}]) \\
    \label{2-2}
    [\boldsymbol{f}^C_{T};\boldsymbol{f}^C_{I}] &= \mathcal{E}_{\text{CLIP}}([\boldsymbol{T}^C_1,...,\boldsymbol{T}^C_N; \boldsymbol{I}^C_{1}, ..., \boldsymbol{I}^C_{M}])
\end{align}

The learnable tokens are used as queries to extract useful information from two models for feature fusion. We discard the output of image tokens to reduce the computational cost. Only the output features corresponding to the learnable tokens, \textit{i.e.}, $\boldsymbol{f}^D_{T}$ and $\boldsymbol{f}^C_{T}$ are fed into the fusion module. 

%It is challenging to effectively fusion features from DINO and CLIP due to the inherent feature-level mismatch. 
Here, we use a bidirectional transformer to fuse the features of DINO and CLIP. The fusion module consists of one cross-attention layer and two self-attention layers. By performing bidirectional cross-attention, the semantic context of CLIP is used to query fine-grained visual details from DINO, and details captured from DINO query the global semantics encoded in CLIP. Specifically, $\boldsymbol{f}^D_{T}$ and $\boldsymbol{f}^C_{T}$ are alternately used as the query, while the other serves as the key and value, as shown in Equation \ref{2-3} and \ref{2-4}. Then, we can obtain two output features, \textit{i.e.,} $\mathbf{H}_{D \to C}$ and $\mathbf{H}_{C \to D}$. 
$\mathbf{H}_{D \to C}$ and $\mathbf{H}_{C \to D}$ are two enhanced features obtained through bidirectional interaction between DINO and CLIP.

\begin{align}
\label{2-3}
    \mathbf{H}_{D \to C}
    &= \text{Transformer}\big(
    \mathbf{Q}=\boldsymbol{f}^D_{T},
    \mathbf{K}=\boldsymbol{f}^C_{T},
    \mathbf{V}=\boldsymbol{f}^C_{T}
    \big) \\
\label{2-4}
    \mathbf{H}_{C \to D}
    &= \text{Transformer}\big(
    \mathbf{Q}=\boldsymbol{f}^C_{T},
    \mathbf{K}=\boldsymbol{f}^D_{T},
    \mathbf{V}=\boldsymbol{f}^D_{T}
    \big)
\end{align}

The outputs obtained from the two queries are concatenated and used as the input to the simple linear classifier $f_{linear}$. This can be formulated as follows:
\begin{equation}
    \label{cls}
    \hat y = f_{linear}([\mathbf{H}_{D \to C};\mathbf{H}_{C \to D}])
\end{equation}

Ultimately, we utilize the standard ReID
loss, \textit{i.e.}, the triplet loss $\mathcal{L}_{\text{Tri}}$ and identity classification loss $\mathcal{L}_{\text{ID}}$ \cite{DBLP:conf/mm/HeLLLCM23}, to optimize our network. The two losses are calculated as follows:

\begin{align}
    \mathcal{L}_{ID} &= \sum_{j=1}^{K} - q_j \log(p_j) \\
    \mathcal{L}_{Tri}& = \max \left( d_p - d_n + \alpha, \, 0 \right)
\end{align}

where $q_j$ denotes the value in the target distribution, $p_j$ represents the ID prediction logits of class $j$, $d_p$ and $d_n$ are the feature distances of the positive and negative pairs, and $\alpha$ is the margin of $\mathcal{L}_{Tri}$.

\subsection{The Intra-model Token Diversity Regularizer}
The intra-model regularizer aims to enable different learnable tokens to capture diverse representations. First, we use side information embeddings (SIE) \cite{DBLP:conf/iccv/He0WW0021} to incorporate camera information into the first learnable token. We add the camera-information embeddings $\boldsymbol{E}_{cam}$ to the first learnable token of DINO and CLIP, respectively, as shown in Equation \ref{3-1} and \ref{3-2}.

\begin{align}
\label{3-1}
    \boldsymbol{T}^D_1 &= \boldsymbol{T}^D_1 + \boldsymbol{E}_{cam} \\
\label{3-2}
    \boldsymbol{T}^C_1 &= \boldsymbol{T}^C_1 + \boldsymbol{E}_{cam}
\end{align}

Then, we maximize the cosine distance, \textit{i.e.}, minimize the cosine similarity, between the first token and vanilla learnable tokens, as shown in Equation \ref{3-3} and \ref{3-4}. This encourages the tokens to capture diverse and complementary information. The first token can thus aggregate more comprehensive features guided by camera information, while the remaining tokens are prompted to learn viewpoint-invariant representations.

\begin{align}
\label{3-3}
    \mathcal{L}^D_{cos} &= \frac{1}{N-1}\sum_{i=2}^N cos(\boldsymbol{f}^D_1, \boldsymbol{f}^D_i) \\
\label{3-4}
    \mathcal{L}^C_{cos} &= \frac{1}{N-1}\sum_{i=2}^N cos(\boldsymbol{f}^C_1, \boldsymbol{f}^C_i) \\
    \mathcal{L}_{intra} &=  \mathcal{L}^D_{cos} + \mathcal{L}^C_{cos}
\end{align}

% \textcolor{red}{With the intra-model regularizer, different learnable tokens can extract complementary features. As shown in Figure \ref{observation} (b), the second token is enabled to pay attention to the backpack, which is not captured by the first token. Figure \ref{observation} (c) indicates that the regularizer further improves performance by 2.5\% on DukeMTMC, demonstrating its effectiveness.}

\subsection{The Inter-model Bias Regularizer}
The inter-model regularizer aims to find the optimal contribution ratio between DINO and CLIP. We have just concatenated the two outputs of the fusion module, \textit{i.e.}, $\mathbf{H}_{D \to C}$ and $\mathbf{H}_{C \to D}$, and input them into the classifier, as shown in Equation \ref{cls}. The linear classifier can be formulated as: 

\begin{align}
\boldsymbol{z}_f&=[\mathbf{H}_{D \to C}; \mathbf{H}_{C \to D}]\\
f_{linear}&=\boldsymbol{W}\boldsymbol{z}_f+ \boldsymbol{b}
\label{classifier}
\end{align}

Equation \ref{classifier} can be further expanded to yield equations as follows:

\begin{align}
f_{linear}&=\boldsymbol{W}_D \cdot \mathbf{H}_{D \to C} + \boldsymbol{b}_D + \boldsymbol{W}_C \cdot \mathbf{H}_{C \to D}+ \boldsymbol{b}_C \\
\boldsymbol{s}^{D} &= \boldsymbol{W}_D \cdot \mathbf{H}_{D \to C} + \boldsymbol{b}_D \\
\boldsymbol{s}^{C} &= \boldsymbol{W}_C \cdot \mathbf{H}_{C \to D}+ \boldsymbol{b}_C
\label{expanding}
\end{align}

where $\boldsymbol{s}^{D}$ denotes the logit output of the DINO and $\boldsymbol{s}^{C}$ denotes the logit output of CLIP. As a result, the final output is the summation of $\boldsymbol{s}^D$ and $\boldsymbol{s}^C$. Thus, the optimization gradient is determined by $\boldsymbol{s}^D$ and $\boldsymbol{s}^C$.

Assume the contribution ratios of the two model outputs are $w_0$ and $w_1$, respectively, and $w_0 + w_1 = 1$. The optimal solution should lead to the minimization of the generalization error $\mathcal{L}$, as shown in Equation \ref{3-7}. 

\begin{align}
    \label{3-7}
    \min_{w_0, w_1} \; g &= \mathbb{E}\mathcal{L}\bigl(f(s), y\bigr) \\
    f(s) & = w_0s^D +w_1 s^C
\end{align}

By performing a bias-variance decomposition of the generalization error \cite{DBLP:journals/corr/abs-1908-05287}, the solution can be rewritten as follows:
\begin{align}
    \label{3-8}
    g  &= (Bias(f(s), y))^2 + Var(f(s)) + Var(\epsilon) \\
    \label{3-5}
    w_0 &= \frac{Bias(s^{C}, y)}{Bias(s^{C}, y) - Bias(s^{D}, y)} \\
    \label{3-6}
    w_1 &= \frac{-Bias(s^{D}, y)}{Bias(s^{C}, y) - Bias(s^{D}, y)}
\end{align}

where $Bias(.)$ is the prediction bias and $Var(.)$ is the variance of $f(s)$. $Bias(s^{D}, y) = \mathbb{E}[s^{D}-y]$ and $Bias(s^{C}, y) = \mathbb{E}[s^{C}-y]$ measures the prediction error of DINO and CLIP, respectively. However, $w_0$ or $w_1$ must be smaller than $0$ in Equation \ref{3-5} and \ref{3-6}, conflicting with $w_0 > 0$ and $w_1 > 0$. Therefore, we cannot find the numerical solution of the equation with the constraint $w_0 + w_1 = 1$. More details of the theoretical derivation can be found in Appendix \ref{A}.

In Equation \ref{3-8}, since we cannot rely solely on re-weighting to minimize the error term $(Bias(f(s), y))^2$, the most effective alternative is to minimize the intrinsic prediction bias of each individual model, \textit{i.e.}, $Bias(s^{D}, y)$ and $Bias(s^{C}, y)$.
This motivates the following proposition:

\begin{proposition}
     \label{lem}
    Given a weighted fusion $f(s) = w_0s^D +w_1 s^C$ with $w_0, w_1 > 0$ and $w_0 + w_1 = 1$, minimizing the generalization error requires minimizing the prediction bias of individual branches.
\end{proposition}

Therefore, maintaining the complementary discriminative strengths of DINO and CLIP, \textit{i.e.}, local detail modeling and global semantic understanding, respectively, is essential for achieving an effective fusion.
We introduce a regularization term for each model to achieve this goal. Specifically, we copy and then feed $\mathbf{H}_{D \to C}$ and $\mathbf{H}_{C \to D}$ into the weight-sharing classifier $f_{linear}$ separately, where the feature duplication is used to align input dimensions.
We use the identity classification loss as the regularization loss:

\begin{align}
    \hat y_D &= f_{linear}([\mathbf{H}_{D \to C}, \mathbf{H}_{D \to C}]) \\
    \hat y_C &= f_{linear}([\mathbf{H}_{C \to D}, \mathbf{H}_{C \to D}])\\
    \mathcal{L}_{inter}&=\mathcal{L}_{ID\_D} + \mathcal{L}_{ID\_C}
\end{align}

% \textcolor{red}{Figure \ref{observation} (e) and (f) presents the experimental results after incorporating the inter-model regularizer. The regularizer effectively eliminates the accuracy gap between DINO and CLIP, and improves performance by a large margin on Market-1501.}

\subsection{Training Loss}

The objective function can be formulated as follows, consisting of the ReID loss, the intra-model regularization loss, and the inter-model regularization loss: 

\begin{equation}
\label{loss}
    L_{total}=L_{ID} + L_{Tri} + \lambda_1 L_{inter} + \lambda_2 L_{intra}
\end{equation}

where $\lambda_1$ and $\lambda_2$ are hyperparameters.

\section{Experiments}
\subsection{Experimental Setup}

\textbf{Datasets and Evaluation Protocols.}
In this study, we evaluate our method on five person re-identification datasets, including Market-1501 \cite{DBLP:conf/iccv/ZhengSTWWT15}, MSMT17 \cite{DBLP:conf/cvpr/WeiZ0018}, DukeMTMC \cite{DBLP:conf/eccv/RistaniSZCT16}, CUHK03-NP \cite{DBLP:conf/cvpr/LiZXW14} and Occluded-Duke \cite{DBLP:conf/iccv/MiaoWLD019}. The details of these datasets are summarized in Table \ref{dataset}. Following existing ReID settings, we use the mean average precision (mAP) and Rank-1 (R-1) accuracy to evaluate the performance.

\begin{table}[t]
  \caption{The statistics of datasets in our experiments.}
  \label{dataset}
  \begin{center}
    \begin{small}
        \begin{tabular}{cccc}
          \toprule
           Dataset & \#ID & Images & Cams \\
          \midrule
           Market-1501 & 1501 & 32668 & 6 \\
           MSMT17 & 4101 & 126441 & 15 \\
           DukeMTMC & 1404 & 36411 & 8 \\
           CHUK03-NP & 1467 & 13164 & 2 \\
           Occuled-Duke & 1404 & 35489 & 8\\
          \bottomrule
        \end{tabular}
    \end{small}
  \end{center}
  \vskip -0.1in
\end{table}

\textbf{Implementation Details.}
We adopt the visual encoders from DINOv2 \cite{DBLP:journals/tmlr/OquabDMVSKFHMEA24} and CLIP \cite{DBLP:journals/corr/abs-2103-00020} as the backbone. We choose the ViT-B/14 for the DINO encoder and ViT-B/16 for the CLIP encoder. Both of them contain 12 transformer layers with the hidden sizes of $768$ dimensions. All input images are resized to $252 \times 126$ for DINO and $256 \times 128$ for CLIP. The number of learnable tokens is set to 2 by default. These tokens are randomly initialized with a dimension of $\mathbb{R}^{N \times 768}$. The fusion transformer consists of one cross-attention layer and two self-attention layers. The final classifier is a simple linear layer that maps the features to the classification labels. We use the Adam optimizer with a learning rate of 5e-6 and consistently set $\lambda_1 = 0.5$ and $\lambda_2 = 5$ in equation \ref{loss} for all datasets. Since the cosine distance loss yields values in the range of $[0,1]$, the coefficient $\lambda_2$ is introduced to rescale it to the same magnitude as other loss terms. The model is trained for 70 epochs. The entire framework is implemented using PyTorch and runs on a single NVIDIA RTX4090 GPU with 24GB VRAM.
%The batch size is configured to 64., with a learning rate warm-up for the early 10 epochs

%An equal number of learnable tokens are used as queries and fed into two encoders together with the image tokens, respectively.

\textbf{Baseline.} The baseline builds on the pre-trained CLIP, and its visual encoder is fine-tuned via the two standard losses, \textit{i.e.}, the triplet loss and identity classification loss.

\subsection{Comparison with State-of-the-Art Methods}

\begin{table*}[t]
  \caption{Comparison with state-of-the-art ViT-based methods on Market-1501, MSMT17, DukeMTMC, CUHK03-NP(labeled) and Occluded-Duke datasets. The best performance is highlighted in \textbf{bold}, and the second-best result is \underline{underlined}.}
  \label{sota}
  \begin{center}
    \begin{small}
        \begin{tabular}{l|c|cccccccccc}
          \toprule
           \multicolumn{1}{c}{\multirow{2}{*}{Methods}} & \multicolumn{1}{|c}{\multirow{2}{*}{References}} &
          \multicolumn{2}{|c}{Market-1501} & 
          \multicolumn{2}{c}{MSMT17} & 
          \multicolumn{2}{c}{DukeMTMC} &
          \multicolumn{2}{c}{CUHK03-NP} &
          \multicolumn{2}{c}{Occluded-Duke} \\
           & & mAP & R-1 & mAP & R-1 & mAP & R-1 & mAP & R-1 & mAP & R-1 \\
          \midrule
           TransReID \cite{DBLP:conf/iccv/He0WW0021} & ICCV'21 & 88.9 & 95.2 & 67.4 & 85.3 & 82.0 & 90.7 & 79.6 & 81.7 & 59.2 & 66.4 \\
           DINO \cite{DBLP:conf/iccv/CaronTMJMBJ21} & ICCV'21 & 90.3 & 95.4 & 64.2 & 83.4 & - & - & - & - & - & - \\
           DINO$+$CFS \cite{DBLP:journals/corr/abs-2111-12084} & Arxiv'21 & 91.0 & 96.0 & 66.1 & 84.6 & - & - & - & - & - & - \\
           DCAL \cite{DBLP:conf/cvpr/ZhuKLLTS22} & CVPR'22 & 87.5 & 94.7 & 64.0 & 83.1 & 80.1 & 89.0 & - & - & - & - \\
           DC-Former \cite{DBLP:conf/aaai/LiZWX0ZCC23} & AAAI'23 & 90.6 & 96.0 & 70.7 & 86.9 & - & - & 79.4 & 81.6 & - & - \\
           CLIP-ReID \cite{DBLP:conf/aaai/LiSL23} & AAAI'23 & 89.6 & 95.5 & 73.4 & 88.7 & 82.5 & 90.0 & 81.6 & 80.9 & \underline{59.5} & \underline{67.1} \\
           AAformer \cite{DBLP:journals/tnn/ZhuGZWLWT24} & TNNLS'23 & 88.0 & 95.4 & 65.6 & 84.4 & 80.9 & 90.1 & 79.0 & 80.3 & - & -\\
           PHA \cite{DBLP:conf/cvpr/ZhangZZ0P23} & CVPR'23 & 90.2 & \underline{96.1} & 68.9 & 86.1 & - & - & \underline{83.0} & \underline{84.5} & - & - \\
           CLIP3DReID \cite{DBLP:conf/cvpr/0037KR024} & CVPR'24 & 88.4 & 95.6 & 61.2 & 81.5 & - & - & - & - & - & - \\
           %CLIMB-ReID \cite{DBLP:conf/aaai/YuLZWZL25}  & AAAI'25 & 92.6 & 96.8 & 77.8 & 90.5 &  - & - & - & - & - & - \\
           FlexiReID \cite{DBLP:conf/icml/SunTSC0DCJ25} & ICML'25 & \underline{92.1} & 96.0 & 67.5 & 83.7 & - & - & - & - & - & - \\
           HPL \cite{zhou2025hierarchicalpromptlearningimage} & AAAI'26 & 89.8 & 95.9 & \textbf{79.0} & \underline{91.0} & \underline{82.9} & \underline{90.3} & - & - & - & - \\
           baseline &  & 86.4 & 93.3 & 66.1 & 84.4 & 80.0 & 88.8 & 80.0 & 80.5 & 53.5 & 60.8 \\
           \textbf{DRFormer} &  & \textbf{92.9} & \textbf{97.0} & \underline{78.7} & \textbf{91.3} & \textbf{85.5} & \textbf{92.5} & \textbf{88.1} & \textbf{89.6} & \textbf{65.3} & \textbf{72.1} \\
          \bottomrule
        \end{tabular}
    \end{small}
  \end{center}
  \vskip -0.1in
\end{table*}

Table \ref{sota} summarizes the comparative results with ViT-based methods. DRFormer outperforms the second-best method and the baseline by a large margin on each benchmark. In particular, our method achieves 85.5\% mAP and 92.5\% R-1 on DukeMTMC, which are 2.6\% and 2.2\% higher than those of the HPL \cite{zhou2025hierarchicalpromptlearningimage} method. On Market-1501 and MSMT17 datasets, our method outperforms FlexiReID \cite{DBLP:conf/icml/SunTSC0DCJ25} by 0.8\% and 11.2\% in terms of mAP, respectively. Compared
with PHA \cite{DBLP:conf/cvpr/ZhangZZ0P23}, DRFormer outperforms it by 5.1\%/5.1\% mAP/Rank-1 on CUHK03-NP dataset. The experimental results demonstrate the effectiveness of our method on five datasets and highlight its strong performance.

\subsection{Ablation Study}
\textbf{Contributions of Different Components.} As shown in Table \ref{ablation}, we conduct experiments on Market-1501 and DukeMTMC to study the effect of each component. `Concat' refers to directly concatenating CLIP and DINO outputs, whereas `Transformer' denotes feature concatenation after bidirectional transformer fusion. The experimental results show that incorporating DINO yields notable gains over the baseline, improving mAP by 3.1\% on Market-1501 and 1.6\% on DukeMTMC. This indicates that supplementing CLIP features with DINO’s fine-grained representations is particularly important. However, directly concatenating the features of the two results in inferior performance. Fusion transformer improves the performance by 1.6\% and 1.2\% in mAP on Market-1501 and DukeMTMC, respectively. The result demonstrates that the fusion module can reduce the discrepancy between DINO- and CLIP-extracted features while preserving their respective strengths. 
%This suggests that the differences in pre-training paradigms and datasets between DINO and CLIP lead to substantial discrepancies in the features they capture.

\begin{table}[t]
  \caption{Ablation of the bidirectional transformer on Market-1501 dataset.}
  \label{ablation}
  \begin{center}
    \begin{small}
        \begin{tabular}{l|cccc}
          \toprule
           \multicolumn{1}{c}{\multirow{2}{*}{Methods}} & \multicolumn{2}{|c}{Market-1501} & \multicolumn{2}{c}{DukeMTMC}\\
           & mAP & R-1 & mAP & R-1\\
          \midrule
           baseline & 86.4 & 93.3 & 80.0 & 88.8 \\
           \ \ $+$Concat & 89.5 & 94.7 & 81.6 & 90.6 \\
           \ \ \ \ \textbf{$+$Transformer} & \textbf{91.1} & \textbf{96.4} & \textbf{82.8} & \textbf{91.0} \\
          \bottomrule
        \end{tabular}
    \end{small}
  \end{center}
  \vskip -0.1in
\end{table}

\textbf{Ablation Study on Regularizers.} To investigate the impact of the inter- and intra-model regularizers, we conduct ablation experiments on Market-1501 and DukeMTMC, as shown in Table \ref{regularizer}. Here, \textit{Intra-model} denotes the intra-model regularizer, and \textit{Inter-model} denotes the inter-model regularizer. Interestingly, on Market-1501, the inter-model regularizer contributes more to performance, improving mAP by 1.4\%, whereas on DukeMTMC, the intra-model regularizer is more effective, yielding a 1.7\% increase in mAP. Applying two regularizers leads to optimal results on both datasets, with improvement of 1.8\%/0.6\% and 2.7\%/1.4\% in mAP/R-1 on Market-1501 and DukeMTMC, respectively. This demonstrates the effectiveness of the two regularizers. The intra-model regularizer enhances the diversity of the features captured by the learnable tokens, and the inter-model regularizer enables a better contribution balance between DINO and CLIP.

\begin{table}[t]
  \caption{Ablation study on the intra- and inter-model regularizers on DukeMTMC dataset.}
  \label{regularizer}
  \begin{center}
    \begin{small}
        \begin{tabular}{cc|cccc}
          \toprule
          \multicolumn{1}{c}{\multirow{2}{*}{Intra-model}} & \multicolumn{1}{c}{\multirow{2}{*}{Inter-model}} & \multicolumn{2}{|c}{Market-1501} & \multicolumn{2}{c}{DukeMTMC} \\
          & & mAP & R-1 & mAP & R-1 \\
          \midrule
          - & - & 91.1 & 96.4 & 82.8 & 91.0 \\
           \checkmark & -  & 91.5 & 96.3 & 84.5 & 92.3 \\
           - & \checkmark & 92.5 & 96.8 & 83.0 & 91.0 \\
          \checkmark & \checkmark  & \textbf{92.9} & \textbf{97.0} & \textbf{85.5} & \textbf{92.5} \\
          \bottomrule
        \end{tabular}
    \end{small}
  \end{center}
  \vskip -0.1in
\end{table}

\textbf{Ablation Study on Fine-tuned Models.} To compare the individual fine-tuning of DINO \cite{DBLP:conf/iccv/CaronTMJMBJ21} and CLIP \cite{DBLP:conf/aaai/LiSL23} with our method, we conducted experiments on Market-1501 and MSMT17. As shown in Table \ref{ft_model}, DINO performs better on Market-1501, whereas CLIP yields superior results on MSMT17. In contrast, our method demonstrates strong performance on both datasets. Our method outperforms DINO by 2.3\% on Market-1501. On the large-scale MSMT17 dataset, our method significantly outperforms the two fine-tuned models, achieving improvements of 14.5\% over DINO and 12.6\% over CLIP. The experimental results demonstrate the effectiveness of our method in capturing and fusing the complementary features of DINO and CLIP.

\begin{table}[t]
  \caption{Experiments of different fine-tuned models on Market-1501 and MSMT17 datasets.}
  \label{ft_model}
  \begin{center}
    \begin{small}
        \begin{tabular}{l|cccc}
          \toprule
           \multicolumn{1}{c}{\multirow{2}{*}{Methods}} & \multicolumn{2}{|c}{Market-1501} & \multicolumn{2}{c}{MSMT17} \\
           & mAP & R-1 & mAP & R-1 \\
          \midrule
           DINO Fine-tuning& 90.3 & 95.4 & 64.2 & 83.4 \\
           CLIP Fine-tuning & 86.4 & 93.3 & 66.1 & 84.4 \\
           \textbf{DRFormer} & \textbf{92.9} & \textbf{97.0} & \textbf{78.7} & \textbf{91.3} \\
          \bottomrule
        \end{tabular}
    \end{small}
  \end{center}
  \vskip -0.1in
\end{table}

\subsection{Hyper-Parameter Analysis}
\textbf{Different Structures of Bidirectional Transformer.} Table \ref{cross_layer} summarizes the experimental results of different designs of the fusion module. As shown in Table \ref{cross_layer}, introducing a single cross-attention layer shows a notable performance improvement. With more self-attention layers, performances can be stably improved. The results demonstrate that performing the cross-attention operation leads to improved performance by facilitating the fusion of the original DINO and CLIP representations.

\begin{table}[t]
  \caption{The impact of various interaction modules on Market-
1501. Cross and self mean cross-attention and self-attention, re-
spectively.}
  \label{cross_layer}
  \begin{center}
    \begin{small}
        \begin{tabular}{l|cc}
          \toprule
           \multicolumn{1}{c|}{Fusion Transformer} & mAP & R-1 \\
          \midrule
           w/o Fusion Module & 89.5 & 94.7 \\
          % \textcolor{red}{$+$1 self-layer} & 91.0 & 95.8 \\
           $+$1 cross-layer & 91.1 & 96.1 \\
           $+$1 cross-layer \& 1 self-layer & 92.0 & 96.8 \\
           \textbf{$+$1 cross-layer \& 2 self-layer} & \textbf{92.9} & \textbf{97.0} \\
          \bottomrule
        \end{tabular}
    \end{small}
  \end{center}
  \vskip -0.1in
\end{table}

\begin{figure}[ht]
  \vskip 0.2in
  \begin{center}
  \centerline{\includegraphics[width=\columnwidth]{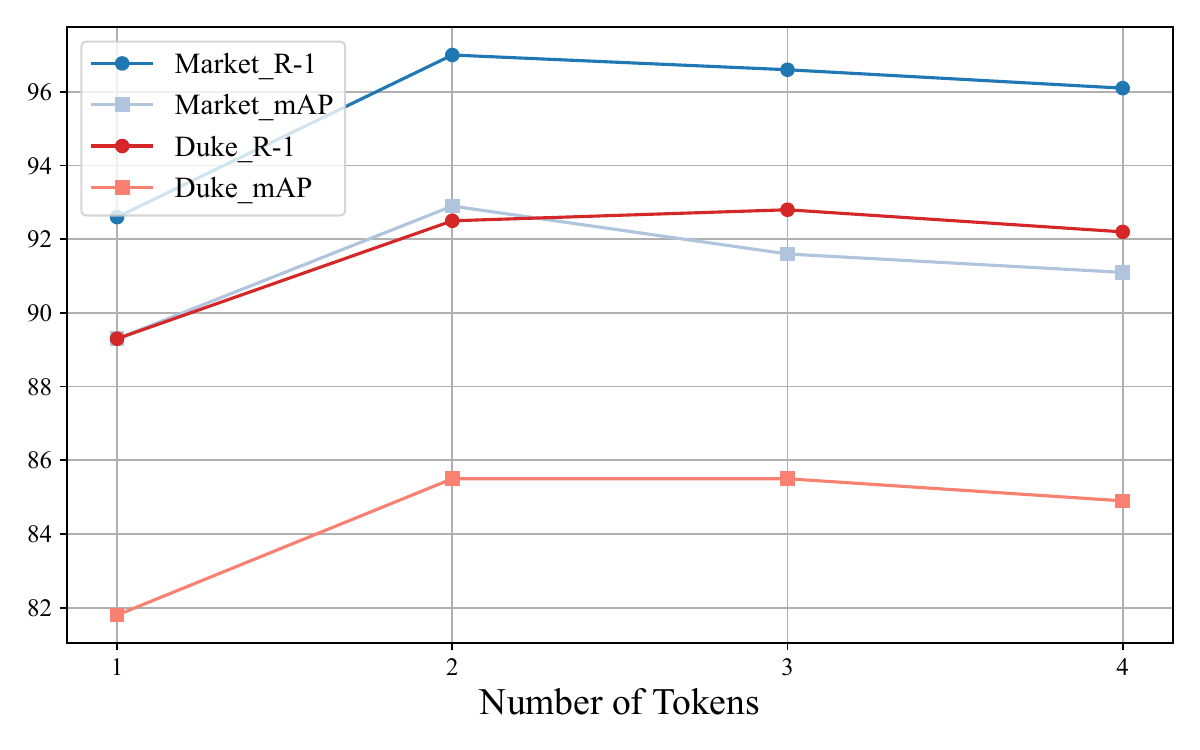}}
    \caption{Visualization of the experiment results with different numbers of learnable tokens on Market-1501 and DukeMTMC datasets.}
    \label{query}
  \end{center}
\end{figure}

\textbf{Different Numbers of Learnable Tokens.} As shown in Figure \ref{query}, different datasets exhibit varying sensitivity to the number of learnable tokens. Generally, only one learnable token may limit the representation capability of the model, leading to suboptimal performance. On Market-1501 and DukeMTMC, the optimal performance is achieved with 2 and 3 learnable tokens, respectively. The above results indicate that we should consider an appropriate number of tokens. Setting too many tokens is not necessarily better.

\section{Qualitative Analysis}
\textbf{Qualitative Results of DRFormer.} Figure \ref{intro} (c) visualizes the attention maps of our method, showing its attention to the holistic body of the person. And Figure \ref{intro} (f) shows that our method enlarges the feature distances between hard samples. These results demonstrate that our approach effectively integrates the capabilities of DINO and CLIP.

\begin{figure}[ht]
  \vskip 0.2in
  \begin{center}
  \centerline{\includegraphics[width=\columnwidth]{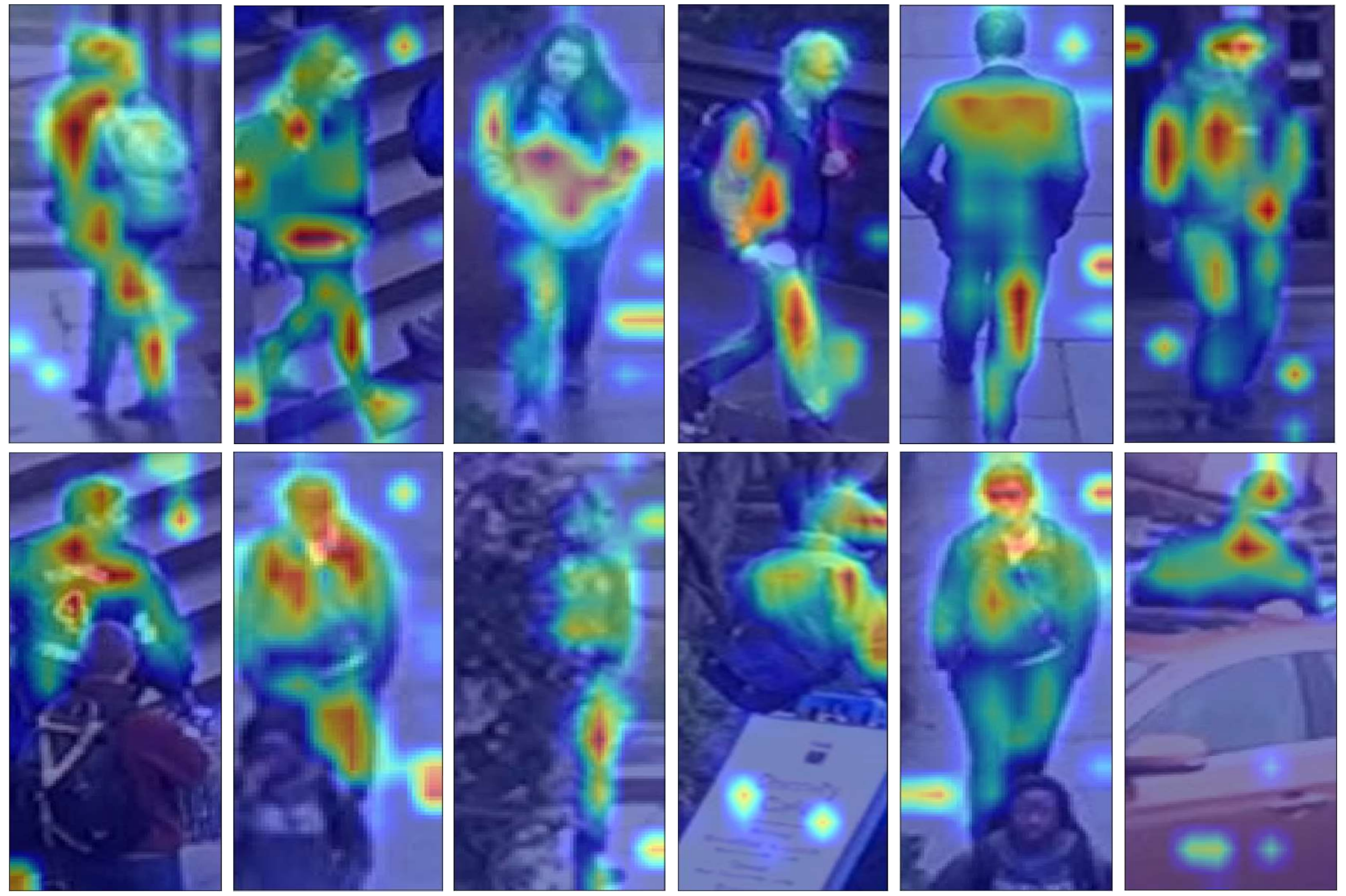}}
    \caption{Visualization of attention weights on DukeMTMC and Occluded-Duke datasets.}
    \label{attn}
  \end{center}
\end{figure}

\textbf{Visualization of Attention Maps.} Figure \ref{attn} visualizes the attention maps of our model on DukeMTMC and Occluded-Duke. Specifically, the first row shows images containing a single pedestrian, while the second row shows images with multiple individuals or complex backgrounds. As shown in Figure \ref{attn}, our model can effectively focus on the human bodies and accurately localize the target pedestrians. The attention maps of our model exhibit clear boundaries with respect to the occlusions and other individuals. 

\section{Conclusion}
In this paper, we analyze and validate the complementary roles of DINO and CLIP in person ReID. DINO effectively captures fine-grained details, while CLIP provides global semantics. To leverage their strengths, we propose a bidirectional transformer module and introduce two regularizers: an intra-model regularizer that encourages diverse feature extraction within each model, and an inter-model regularizer that balances the contributions of DINO and CLIP by theoretical guidance.
Extensive quantitative and qualitative experiments confirm the effectiveness of DRFormer.
%This mechanism can be naturally extended to other tasks.
%Such a mechanism is not limited to person ReID but can also be applied to other tasks that require both strong generalization and fine-grained discrimination.

%that demand strong generalization and fine-grained discrimination.

%We believe that effectively combining the fine-grained features of DINO with the global semantics of CLIP is challenging but essential. 
%The intra-model regularizer promotes the feature diversity, and the inter-model regularizer achieves a better contribution balance of the two models.
%
\section*{Impact Statement}
This paper presents work whose goal is to advance the field of Machine Learning. There are many potential societal consequences of our work, none
which we feel must be specifically highlighted here.
\bibliography{main}
\bibliographystyle{icml2026}

%%%%%%%%%%%%%%%%%%%%%%%%%%%%%%%%%%%%%%%%%%%%%%%%%%%%%%%%%%%%%%%%%%%%%%%%%%%%%%%
%%%%%%%%%%%%%%%%%%%%%%%%%%%%%%%%%%%%%%%%%%%%%%%%%%%%%%%%%%%%%%%%%%%%%%%%%%%%%%%
% APPENDIX
%%%%%%%%%%%%%%%%%%%%%%%%%%%%%%%%%%%%%%%%%%%%%%%%%%%%%%%%%%%%%%%%%%%%%%%%%%%%%%%
%%%%%%%%%%%%%%%%%%%%%%%%%%%%%%%%%%%%%%%%%%%%%%%%%%%%%%%%%%%%%%%%%%%%%%%%%%%%%%%
\newpage
\appendix
\onecolumn
\section{Theoretical Derivation}
\label{A}
%You can have as much text here as you want. The main body must be at most $8$ pages long. For the final version, one more page can be added. If you want, you can use an appendix like this one.

%The $\mathtt{\backslash onecolumn}$ command above can be kept in place if you prefer a one-column appendix, or can be removed if you prefer a two-column appendix.  Apart from this possible change, the style (font size, spacing, margins, page numbering, etc.) should be kept the same as the main body.
%%%%%%%%%%%%%%%%%%%%%%%%%%%%%%%%%%%%%%%%%%%%%%%%%%%%%%%%%%%%%%%%%%%%%%%%%%%%%%%
%%%%%%%%%%%%%%%%%%%%%%%%%%%%%%%%%%%%%%%%%%%%%%%%%%%%%%%%%%%%%%%%%%%%%%%%%%%%%%%

In Section 3.4, we briefly explain the motivation for introducing the inter-model regularizer. Here, we provide a detailed derivation of the formulation.

\textbf{Dual-Model Learning.} Without loss of generality, we consider two encoders as $\varphi_0(\boldsymbol{\theta}_0, \cdot)$ and $\varphi_1(\boldsymbol{\theta}_1, \cdot)$, where $\theta_0$ and $\theta_1$ are the parameters of encoders. 
The dataset is denoted as $\mathcal{D} = {\{x_i\}}_{i=1,2,...,N}$, where $y \in {1, 2, . . . ,M}$, and $M$ is the number of categories. 
The representation outputs of encoders are denoted as $\boldsymbol{f}_0 = \phi_0(\boldsymbol{\theta}_0, x_i)$ and $\boldsymbol{f}_1 = \phi_1(\boldsymbol{\theta}_1, x_i)$. 
The two encoders are connected through the representations by some kind of fusion method. Here, let $\phi_f(\boldsymbol{\theta}_{\boldsymbol{f}}, \cdot)$ denotes the fusion module. $\boldsymbol{\theta}_f$ is the parameter of this module. Notably, our method employs bidirectional cross-attention for fusion. As a result, the fusion module outputs two enhanced feature representations $\boldsymbol{z}_0$ and $\boldsymbol{z}_1$.

\begin{align}
    \boldsymbol{z}_0 &= \phi_f(\boldsymbol{\theta}_{\boldsymbol{f}}, \boldsymbol{f}_0)\\
    \boldsymbol{z}_1 &= \phi_f(\boldsymbol{\theta}_{\boldsymbol{f}}, \boldsymbol{f}_1)
\end{align}

Let $\boldsymbol{W} \in \mathbb{R}^{M\times(d_0+d_1)}$ and $\boldsymbol{b} \in \mathbb{R}^M$ denote the parameters of the linear classifier to produce the logits output. The output of input $x_i$ in a dual-model can be expressed as follows:

\begin{align}
    f(x_i) &= \boldsymbol{W}\boldsymbol{z}_f + \boldsymbol{b}\\
    \boldsymbol{z}_f & = [\boldsymbol{z}_0; \boldsymbol{z}_1]
\end{align}

Let $\boldsymbol{s}^0_i = \boldsymbol{W}_0 \cdot \boldsymbol{z}_0 +\boldsymbol{b}_0$, which denotes the logit output of model $\varphi_0(\boldsymbol{\theta}_0, \cdot)$, and $\boldsymbol{s}^1_i = \boldsymbol{W}_1 \cdot \boldsymbol{z}_1 + \boldsymbol{b}_1$, which denotes
the logit output of model $\varphi_1(\boldsymbol{\theta}_1, \cdot)$. As a result, the final output is the summation of $\boldsymbol{s}^0_i$ and $\boldsymbol{s}^1_i$. Thus, the gradient is determined by $\boldsymbol{s}^0_i$ and $\boldsymbol{s}^1_i$.

\textbf{The optimal contribution ratio for different models}. We need to find the optimal contribution ratio of different models. Generally, the optimal contribution ratio should minimize the generalization
error, thus we formulate the problem as follows,

\begin{align}
    \min_{w_0,w_1} g &= \frac{1}{N} \sum^{i=N}_{i=1} \mathcal{L}(f(x_i), y_i) \\
    f(x_i) & = w_0s^0_i+w_1s^1_i
\end{align}

where $w_0 > 0$ and $w1 > 0$ denote the contribution of two models, respectively, and $w_0 + w_1 = 1$.
$\mathcal{L}(.)$ measures the generalization error from prediction and ground truth. For simplification, we define $E [f(x)] = \frac{1}{N} \sum^{i=N}_{i=1} f(x_i)$. According to the bias-variance decomposition \cite{DBLP:journals/corr/abs-1908-05287}, the error between prediction and ground truth can be rewritten as follows. Here, we mainly focus on the squared term.

\begin{equation}
    \begin{cases} g  = (Bias(f(x), y))^2 + Var(f(x)) + Var(\epsilon) \\
    Bias(f(x_i), y) = \mathbb{E}[f(x)-y]
    \end{cases}
\end{equation}

Then, with the constraint $w_0 +w_1 = 1$, we get the solution of $w_0$ and $w_1$ as follows,

\begin{align}
    w_0 &= \frac{Bias(s^{0}, y)}{Bias(s^{0}, y) - Bias(s^{1}, y)} \\
    w_1 &= \frac{-Bias(s^{1}, y)}{Bias(s^{0}, y) - Bias(s^{1}, y)}
\end{align}

Here, the numerical solution is meaningless since one of $w_0$ or $w_1$ must be smaller than 0, conflicting with $w_0 > 0$ and $w_1 > 0$. Thus, when $Bias(s^0 , y)$ and $Bias(s^1 , y)$ are fixed, we can not find a reasonable combination of $w0$ or $w1$ to minimize $Bias(.)^2$. Consequently, the only way to minimize $Bias(.)^2$ is to minimize $Bias(s^0 , y)$ and $Bias(s^1 , y)$.

\section{Visualization of Classification Results}
\begin{figure}[ht]
  \vskip 0.2in
  \begin{center}
  \centerline{\includegraphics[width=0.8\columnwidth]{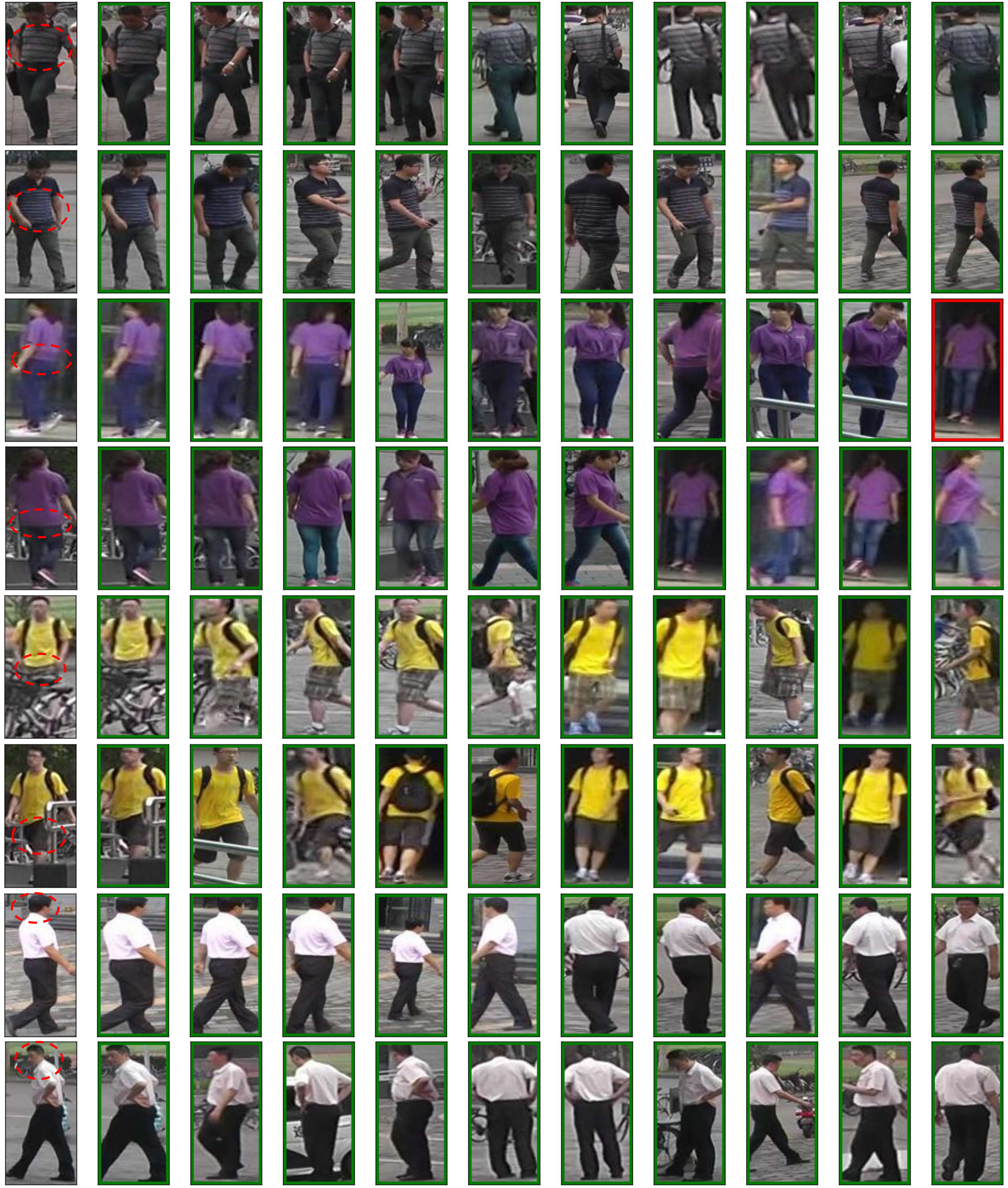}}
    \caption{Classification results of highly similar samples on Market-1501.}
    \label{rank}
  \end{center}
\end{figure}

\textbf{Hard Pedestrian Sample Discrimination.} Figure \ref{rank} illustrates several pedestrian images with different identities, but highly similar appearances and body poses. The red dashed lines mark the differences in fine-grained details across images of different pedestrians. Our model successfully distinguished these challenging samples, demonstrating its strong capability for image understanding.
%Therefore, it is key for the model to acquire strong detail understanding capability while accurately localizing the target.

\section{Ablation Study on Intra-model Regularizer.}

\textbf{Ablation on Components of the Intra-model Regularizer.} The intra-model regularizer first applies SIE and then maximizes the cosine distance between tokens. We conduct ablation studies on these two components on the DukeMTMC and Occluded-Duke datasets. As shown in Table \ref{intra}, SIE improves the mAP by 1.2\% and 0.3\% on DukeMTMC and Occluded-Duke, respectively. The cosine loss further boosts the performance by 1.3\% and 1.0\%. Overall, the cosine loss has a greater impact on performance. The experiment results validate the effectiveness of the intra-model regularizer.

\begin{table}[t]
  \caption{Ablation on components of the intra-model regularizer on DukeMTMC and Occluded-Duke datasets.}
  \label{intra}
  \begin{center}
    \begin{small}
        \begin{tabular}{l|cccccc}
          \toprule
           \multicolumn{1}{c}{\multirow{2}{*}{Methods}} & \multicolumn{2}{|c}{DukeMTMC} & \multicolumn{2}{c}{Occluded-Duke} \\
           & mAP & R-1 & mAP & R-1 \\
          \midrule
           w/o Intra-model & 83.0 & 91.0 & 64.0 & 72.4 \\
           \ \ $+$SIE & 84.2 & 91.5 & 64.3 & 72.1 \\
           \ \ \ \ $+$Cosine Loss & 85.5 & 92.5 & 65.3 & 72.3 \\
          \bottomrule
        \end{tabular}
    \end{small}
  \end{center}
  \vskip -0.1in
\end{table}

\end{document}